\documentclass{article}

\usepackage[preprint]{neurips_2026}

\usepackage[utf8]{inputenc}
\usepackage[T1]{fontenc}
\usepackage{hyperref}
\usepackage{url}
\usepackage{booktabs}
\usepackage{amsfonts}
\usepackage{amsmath}
\usepackage{amssymb}
\usepackage{nicefrac}
\usepackage{microtype}
\usepackage{xcolor}
\usepackage{colortbl}
\usepackage{graphicx}
\usepackage{tabularx}
\usepackage{array}
\usepackage{multirow}
\usepackage{algorithm}
\usepackage{algpseudocode}
\usepackage[font=small,labelfont=bf]{caption}

\newcommand{\coloredhref}[3][blue]{%
  \href{#2}{\textcolor{#1}{#3}}%
}
\definecolor{burgundy}{rgb}{0.5, 0.0, 0.13}

\newcommand{\method}{\textsc{Prune-OPD}}
\newcommand{\opd}{\textsc{OPD}}
\newcommand{\grpo}{\textsc{GRPO}}

\newcommand{\student}{\pi_{\theta}}
\newcommand{\teacher}{\pi_{\mathrm{T}}}

\newcommand{\accept}{\mathcal{A}_{\tau}}
\newcommand{\topk}{\mathcal{K}_{\tau}}
\newcommand{\overlap}{\mathcal{O}_{\tau}}

\newcommand{\clip}{\mathrm{clip}}

\title{\method: Efficient and Reliable On-Policy Distillation for Long-Horizon Reasoning}

\vspace{-3mm}
\author{
Zhicheng Yang\textsuperscript{1}~~Zhijiang Guo\textsuperscript{1,2}~~Yifan Song\textsuperscript{1}~~Minrui Xu\textsuperscript{1}~~Yongxin Wang\textsuperscript{3}\\
\textbf{Yiwei Wang}\textsuperscript{4}~~\textbf{Xiaodan Liang}\textsuperscript{3,5}~~\textbf{Jing Tang}\textsuperscript{1,2}\\
$^1$The Hong Kong University of Science and Technology (Guangzhou) \\
$^2$The Hong Kong University of Science and Technology \\
$^3$MBZUAI, $^4$University of California, Merced, $^5$Sun Yat-sen University \\
\\
\textit{\textbf{Project Repo}}: \coloredhref[burgundy]{https://github.com/yangzhch6/Prune-OPD}{\ttfamily \textbf{https://github.com/yangzhch6/Prune-OPD}}
}

\begin{document}

\maketitle

\begin{abstract}
On-policy distillation (OPD) leverages dense teacher rewards to enhance reasoning models. However, scaling OPD to long-horizon tasks exposes a critical flaw: as the student's generated prefix inevitably diverges from the teacher's thought process, the teacher's dense reward loses local exploitability. Continuing to generate and evaluate tokens on these ``drifted'' trajectories not only degrades reward quality but also incurs massive computational waste. 
To address this, we introduce \textbf{Prune-OPD}, a framework that dynamically aligns training budgets with supervision quality. By continuously monitoring the local compatibility between student and teacher predictions (e.g., via top-$k$ overlap), Prune-OPD detects prefix-drift events in real time. Upon detecting severe drift, it monotonically down-weights subsequent unreliable rewards and triggers dynamic rollout truncation. This allows the training process to halt futile generation and reallocate compute strictly to reliable teacher supervision.
Across diverse teacher-student combinations, Prune-OPD consistently aligns computation with supervision reliability. When prefix drift makes dense teacher rewards unreliable, it reduces training time by 37.6\%--68.0\% while preserving, and often improving, performance on challenging benchmarks (AMC, AIME, HMMT). When student-teacher compatibility remains high, it automatically preserves long-context supervision by expanding the training window. These results suggest that Prune-OPD improves OPD not by blindly shortening rollouts, but by reallocating computation toward locally exploitable teacher rewards.

\end{abstract}
\vspace{-4mm}

\section{Introduction}
\vspace{-2mm}
On-policy distillation (OPD) has become a central technique for post-training large language models.
Recent systems like Qwen3~\citep{yang2025qwen3}, MiMo~\citep{xiao2026mimo}, and GLM-5~\citep{zeng2026glm} adopt OPD-style dense supervision in post-training pipelines. Thinking Machines Lab~\citep{lu2025onpolicydistillation} reports that an OPD recipe can reproduce strong reasoning gains at substantially lower compute than outcome-reward RL.
This makes OPD an attractive complement to supervised fine-tuning and verifiable RL: it supplies a token-level learning signal on long reasoning traces, rather than waiting for a sparse final-answer reward.

OPD's appeal lies in its on-policy nature. Off-policy distillation trains the student on fixed teacher-generated sequences, which induces the exposure-bias problem: at inference time, the student must act under prefixes that were not produced by the teacher or reference policy~\citep{bengio2015scheduled}.
OPD instead samples rollouts from the current student and evaluates the teacher on the same student-visited prefixes.
This design has also motivated recent self-distillation methods, where a model acts as its own teacher under privileged information or feedback and improves from its own on-policy experience~\citep{hubotter2026reinforcement,zhao2026self,shenfeld2026self}.
In long-horizon mathematical reasoning, this distinction matters: a single solution can contain thousands of tokens, so even small train-inference mismatches can compound across the trajectory.

However, the same on-policy design creates a new reliability problem.
The teacher is queried not only on prefixes where its local distribution offers useful corrections, but also on prefixes that have already drifted away from the teacher's reasoning process.
Recent OPD dynamics analysis points out that OPD success depends on local thinking-pattern compatibility and that reward quality can deteriorate with trajectory depth~\citep{li2026rethinking}.
We take this diagnosis as a design constraint: long-horizon OPD should not treat every generated token as equally worth supervising.

The missing piece is online reliability control.
Current OPD recipes usually choose a rollout budget before generation and then apply dense teacher rewards uniformly along the sampled response.
This creates a bad tradeoff.
A short fixed budget can discard useful long reasoning, while a long fixed budget spends substantial compute generating and scoring suffixes whose teacher signal may no longer be locally exploitable.
In practice, a single rollout can contain both regimes: an early prefix where student and teacher still share plausible next-token support, followed by a suffix where the student has committed to reasoning choices the teacher would not have taken.
What OPD needs is therefore not simply more or less dense reward, but a way to decide \emph{where} the dense reward is reliable enough to train on.

We propose \method, a reliability-aware modifier for long-horizon OPD that makes this decision during training.
Rather than filtering whole responses or imposing a fixed response length, \method~asks a position-level question: under the same student prefix, do the student and teacher still share a high-probability candidate region?
The primary signal is per-position top-$k$ overlap ratio, with teacher top-$p$/top-$k$ action acceptance as a stricter variant.
When the selected compatibility metric falls below its threshold, \method~records a prefix-drift event.
Cumulative drift events define a monotone reliability weight that attenuates later OPD rewards, and the same raw reliability signal defines an effective response length for dynamic rollout control.
The intended shift is from fixed-budget OPD to reliability-budgeted OPD: dense supervision should be allocated where the teacher remains actionable on the student's trajectory.

\begin{figure}[t]
\vspace{-3mm}
\centering
\includegraphics[width=\linewidth]{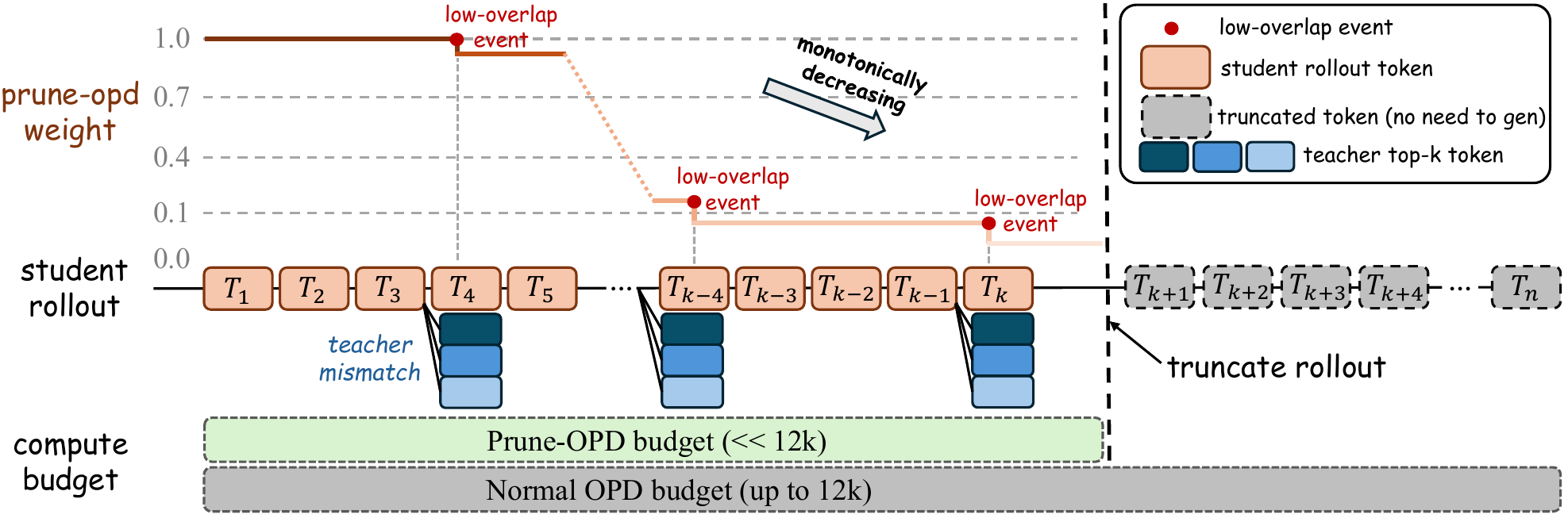}
\vspace{-1mm}
\caption{Conceptual overview of \method. \method~monitors local student-teacher compatibility along the student rollout, monotonically attenuates OPD rewards after low-overlap drift events, and truncates the response once reliable teacher supervision is exhausted.}
\label{fig:overview}
\vspace{-6mm}
\end{figure}

We conduct extensive experiments to evaluate whether reliability-aware pruning improves long-horizon OPD in practice.
The results validate the effectiveness of \method: it reduces training time by 37.6\%--68.0\% when dense supervision becomes unreliable, while preserving or improving downstream accuracy.
Additional runs show that the controller keeps long-context supervision when the reliability signal remains high, supporting the claim that \method~reallocates computation rather than blindly shortening rollouts. In summary, our contributions are:

\vspace{-2mm}
{\setlength{\leftmargini}{1.6em}
\begin{itemize}
    \item We formulate long-horizon OPD as a reliability-allocation problem on student-visited prefixes, connecting trajectory-depth degradation to local compatibility and reward exploitability.
    \item We propose \method, which converts cumulative prefix drift into monotone OPD reward attenuation and a dynamic response-length controller, without changing the baseline OPD path when disabled.
    \item We provide pilot evidence that reliability-aware pruning can reduce OPD training time by 37.6\%--68.0\% when prefix drift limits reliable supervision while preserving benchmark performance, and we specify fixed-length pruning controls to test whether the gains come from reliability rather than generic shortening.
\end{itemize}
}

\section{Background and Problem Setup}

\subsection{On-Policy Distillation}

Let $x$ denote an input prompt and $y=(y_1,\ldots,y_T)$ a response sampled autoregressively from a student policy $\student$.
At position $t$, write $y_{<t}=(y_1,\ldots,y_{t-1})$ for the prefix and define the student and teacher next-token distributions as:
\begin{equation}
    p_t(v) \triangleq \student(v\mid x,y_{<t}), \qquad
    q_t(v) \triangleq \teacher(v\mid x,y_{<t}),
    \label{eq:pt-qt}
\end{equation}
over vocabulary tokens $v\in\mathcal{V}$.
OPD computes supervision on trajectories sampled from the current student, usually by minimizing a reverse-KL objective over student-generated trajectories \citep{gu2024minillm,agarwal2024policy,li2026rethinking}:
\begin{equation}
    \mathcal{L}_{\mathrm{OPD}}(\theta)
    =
    \mathbb{E}_{x,\;y\sim\student(\cdot\mid x)}
    \left[
        \sum_{t=1}^{T}
        D_{\mathrm{KL}}\!\left(p_t \,\|\, q_t\right)
    \right].
    \label{eq:opd-token-kl}
\end{equation}
This token-level decomposition drives OPD's sample efficiency: every position provides a dense teacher signal instead of waiting for a final-answer reward.
It also exposes the central assumption: $q_t$ should be a useful local target on the student-visited state $(x,y_{<t})$.
In practice, we approximate this reward on the student's top-$k$ candidate tokens and obtain a per-position reward vector $r_{t,\cdot}$.
Because rewards are already localized by position, a reliability modifier can simply multiply all rewards at position $t$ by a scalar weight before the policy loss.
This is the interface used by \method: reliable prefixes keep their OPD rewards, while drifted prefixes receive attenuated rewards.

\subsection{Dynamic Metrics}

Following the OPD mechanism analysis~\citep{li2026rethinking}, let $S_t^{(p)}=\operatorname{TopK}(p_t,k)$ and $S_t^{(q)}=\operatorname{TopK}(q_t,k)$.
The overlap ratio proposed in that work is
\begin{equation}
    \mathcal{M}_{\mathrm{overlap}}
    =
    \mathbb{E}_{t}\left[
        \frac{|S_t^{(p)}\cap S_t^{(q)}|}{k}
    \right],
    \label{eq:overlap-ratio}
\end{equation}
which measures whether the student and teacher assign high probability to a shared candidate region.
The entropy gap,
\begin{equation}
    \Delta H_t = |H(q_t)-H(p_t)|,
    \label{eq:entropy-gap}
\end{equation}
tracks whether the two models have similar uncertainty on the same visited state.
These metrics are usually reported as diagnostics after training.
\method~turns them into online control signals.
The current primary metric is the per-position overlap ratio, which measures whether student and teacher still share a high-probability candidate region on the same student prefix.
We also evaluate a stricter action-level variant that asks whether the sampled student token is accepted by the teacher's observable top-$p$/top-$k$ region.

\subsection{Long-Horizon Reliability Problem}

The reward at position $t$ depends on $q_t(v)=\teacher(v\mid x,y_{<t})$, where the prefix $y_{<t}$ was produced by the student.
As $t$ grows, the prefix can drift away from trajectories the teacher would naturally generate.
Prior OPD analysis reports suffix-to-prefix instability and decreasing teacher-continuation advantage as the student prefix depth increases \citep{li2026rethinking}.
Thus the same dense reward that helps on moderate traces can become unreliable on long traces. This motivates position-level reliability control.
We estimate whether each student prefix remains teacher-compatible, then allocate dense supervision to reliable prefixes and attenuate rewards after prefix drift.

\section{\method}
\label{sec:method}

\method~converts local student-teacher compatibility into a position-wise weight on OPD rewards.
The method is designed as a drop-in modifier for OPD: it does not alter how the teacher reward is computed, but changes how much each visited position should trust that reward.
It has three parts: a per-position compatibility metric, a cumulative reliability weight, and an optional dynamic response-length controller.

\subsection{Compatibility Metric}

At position $\tau$, our goal is to decide whether the teacher still provides locally exploitable supervision on the student's current trajectory.
The student and teacher are therefore evaluated on the same student-generated
prefix $s_{\tau}=(x,y_{<\tau})$.

We first restrict the comparison to each model's high-confidence candidate
tokens. Let
\begin{equation}
    \mathcal{K}^{\mathrm{S}}_{\tau}
    =
    \operatorname{TopK}\!\left(\student(\cdot\mid s_{\tau}),k\right),
    \qquad
    \mathcal{K}^{\mathrm{T}}_{\tau}
    =
    \operatorname{TopK}\!\left(\teacher(\cdot\mid s_{\tau}),k\right)
\end{equation}
be the returned top-$k$ candidate sets.
The top-$k$ size is an experimental hyperparameter.
Then, we measure local compatibility with the overlap ratio
\begin{equation}
    \overlap
    =
    \frac{
    \left|\mathcal{K}^{\mathrm{S}}_{\tau}\cap \mathcal{K}^{\mathrm{T}}_{\tau}\right|
    }{k}.
    \label{eq:local-overlap}
\end{equation}
Low overlap means that the teacher's dense reward is being computed under prefixes where its preferred next-token support has separated from the student's support. Accordingly, we define the prefix-drift event as
\begin{equation}
    B_{\tau}
    =
    \mathbf{1}\left[\overlap < \gamma\right],
    \label{eq:bad-event}
\end{equation}
where $\gamma$ is the compatibility threshold.
These drift events are later accumulated to attenuate subsequent OPD rewards and define the effective response length for dynamic rollout control.

This metric follows the OPD mechanism analysis directly: successful OPD is associated with increasing high-probability overlap, while failing runs show stagnant or unstable overlap on student-visited states \citep{li2026rethinking}.
It measures whether the student and teacher still share a plausible local action space.
Low overlap means that the teacher's dense reward is being computed in a region where its preferred next-token support has separated from the student's support.
We also design an alternative top-$p$ action-acceptance metric, its definition is given in Appendix~\ref{app:top-p-metric}.

\subsection{Cumulative Reliability and Loss Weighting}

Bad events are accumulated over the response:
\begin{equation}
    C_{\tau} = \sum_{i=1}^{\tau} B_i.
\end{equation}
The raw reliability weight is a clipped linear decay,
\begin{equation}
    R_{\tau} = \clip(1 - w_{\mathrm{drop}} C_{\tau}, 0, 1),
    \label{eq:raw-weight}
\end{equation}
where $w_{\mathrm{drop}}\geq 0$ controls the penalty per compatibility failure.
The raw weight is monotone non-increasing along each valid response.
Padding positions are forced to zero.

The actual loss weight adds a base floor:
\begin{equation}
    L_{\tau} = R_{\tau} + w_{\mathrm{base}}
    \label{eq:loss-weight}
\end{equation}
for valid response positions, and $L_{\tau}=0$ for padding.
The scaled \opd~reward is
\begin{equation}
    \widetilde{r}_{\tau,j} = L_{\tau} \cdot r_{\tau,j}.
    \label{eq:scaled-reward}
\end{equation}
The base weight has two practical roles.
First, it prevents \method~from becoming a hard switch when $R_{\tau}$ reaches zero.
Second, it lets the method trade off denoising and retention of weak teacher signal.
Concrete hyperparameter values are specified in the experimental setup.

\subsection{Dynamic Response Budget}

\method~can additionally adjust the rollout length during training.
Define the effective response length of a sample as
\begin{equation}
    E = \sum_{\tau=1}^{T} \mathbf{1}\left[R_{\tau}>\epsilon\right],
    \label{eq:effective-length}
\end{equation}
for a small tolerance $\epsilon$.
This definition deliberately uses $R_{\tau}$, not $L_{\tau}$.
Therefore a token with $R_{\tau}=0$ and $w_{\mathrm{base}}>0$ can still contribute a small \opd~reward, but it is not counted as reliable length.

% \vspace{-2mm}
\begin{algorithm}[t]
\caption{\method~with reliability-aware reward scaling and dynamic response budget}
\label{alg:prune-opd}
\begin{algorithmic}[1]
\Require prompt distribution $\mathcal{D}$, student policy $\student$, teacher policy $\teacher$, OPD reward routine, initial response budget $M_1$
% \Require compatibility, reward-weighting, and response-budget hyperparameters
\Ensure updated student policy $\student$ and response budget sequence $\{M_t\}$
\For{training step $t=1,2,\ldots$}
    \State Sample prompts $\{x^{(b)}\}_{b=1}^{B}\sim\mathcal{D}$ and generate student rollouts $\{y^{(b)}_{1:T_b}\}_{b=1}^{B}$ with maximum length $M_t$.
    \State Compute raw OPD reward tensors $r^{(b)}_{\tau,j}$ on the student-visited prefixes.
    \For{each rollout $b=1,\ldots,B$}
        \State Compute the overlap-ratio sequence on student-visited prefixes using Eq.~\eqref{eq:local-overlap}.
        \State Mark low-compatibility positions and form the cumulative reliability weights using Eqs.~\eqref{eq:bad-event}--\eqref{eq:loss-weight}.
        \State Scale OPD rewards using Eq.~\eqref{eq:scaled-reward} and compute the effective reliable length $E^{(b)}$ using Eq.~\eqref{eq:effective-length}.
    \EndFor
    \State Update $\theta$ with the OPD objective using scaled rewards $\widetilde{r}^{(b)}_{\tau,j}$.
    \State $h_t\leftarrow B^{-1}\sum_{b=1}^{B}\mathbf{1}[E^{(b)}\geq M_t-m]$.
    \If{$h_t\geq \rho$}
        \State $M_{t+1}\leftarrow \min(M_t+\Delta,M_{\max})$ and reset the low-hit counter.
    \ElsIf{$h_t<\rho$ for $P$ consecutive steps}
        \State $M_{t+1}\leftarrow \max(M_t-\Delta,M_{\min})$ and reset the low-hit counter.
    \Else
        \State $M_{t+1}\leftarrow M_t$.
    \EndIf
\EndFor
\end{algorithmic}
\end{algorithm}
% \vspace{-8mm}

% \vspace{-5mm}
Let $M_t$ be the maximum response length used for rollout generation at training step $t$.
The controller monitors the batch hit ratio: the fraction of samples whose effective reliable length reaches the current limit up to a margin.
If the hit ratio is large, the limit expands by a fixed step, capped by $M_{\max}$.
If the hit ratio remains small for a patience window, the limit contracts by the same step, bounded below by $M_{\min}$.
The initial limit, bounds, update step, margin, hit-ratio threshold, and patience are specified in the experimental setup. Algorithm~\ref{alg:prune-opd} summarizes the full \method~procedure, including reliability-aware reward scaling and dynamic response-budget adaptation.
To keep the training loop readable, the position-wise compatibility and weighting rules refer directly to the definitions above.

% The implementation is intentionally narrow.
% When \method~is disabled, the original OPD or GRPO path is unchanged.
% When enabled, \method~only rescales the already-computed OPD reward tensor.
% It does not currently route zero-reliability tokens to a GRPO fallback loss; that is an explicit future direction rather than part of the present method.
\section{Experiments}
\label{sec:experiments}

We evaluate whether reliability-aware reward scaling can substantially reduce OPD training cost while preserving downstream reasoning accuracy.
Beyond final benchmark accuracy and training time, we measure local student-teacher dynamics on student-visited states, including overlap ratio, top-$p$ acceptance, and effective reliable length.

\subsection{Experimental Setup}

\noindent\textbf{Training data.}
We train on DAPO-Math-17K \citep{yu2025dapo}, following the OPD setting used in recent reasoning experiments.
Training prompts are shuffled before each run.
The primary RL outcome reward is final-answer correctness. However, OPD supplies dense token-level distillation rewards.

\noindent\textbf{Models.}
We evaluate five teacher-student pairs to test the reliability signal across distinct OPD dynamics.
DeepSeek-R1-Distill-Qwen-1.5B is distilled from DeepSeek-R1-Distill-Qwen-7B, a larger same-family reasoning model whose higher benchmark score may not imply locally exploitable token-level feedback \citep{guo2025deepseek}.
The same student is also distilled from JustRL-DeepSeek-1.5B, a same-size RL-improved teacher obtained from DeepSeek-R1-Distill-Qwen-1.5B \citep{he2025justrl}.
Qwen3-1.7B-Base and Qwen3-4B-Base are distilled from Qwen3-4B (Non-thinking), isolating how a thinking-mode shift interacts with student scale within the same model family \citep{yang2025qwen3}.
Finally, DeepSeek-R1-Distill-Qwen-7B is distilled from Skywork-OR1-7B, providing a high-initial-overlap setting where \method~is expected to preserve long-context supervision rather than shorten the rollout.
Together, these combinations test the same claim under different conditions: the controller reduces computation when reliability decays, while preserving dense supervision when compatibility remains high.

\noindent\textbf{Baselines.}
We compare baseline OPD, fixed-length truncation baselines, and \method~variants.
Fixed-length OPD uses the same distillation objective as baseline OPD but clamps the maximum response length to a fixed budget.
Random pruning matches either the retained token fraction or retained OPD reward mass of \method~while choosing pruned positions independently of student-teacher compatibility.
These controls test whether the benefit comes from the student-teacher compatibility signal rather than from generic shorter rollouts or lower reward density.
The main \method~configuration uses overlap ratio with threshold $\gamma=0.7$.
We also report a teacher top-$p$ action-acceptance variant with $p=0.95$. Both variants use top-$k=16$, $w_{\mathrm{drop}}=0.01$, $w_{\mathrm{base}}=0.5$, and dynamic-length hit ratio $0.1$.
Unless otherwise stated, dynamic response length uses initial 2048, min 1024, max 12288, update step 100, margin 100, and shrink patience 3.

\noindent\textbf{Evaluation.}
We evaluate on AMC23, AIME24, AIME25, HMMT24, and HMMT25.
The primary metric is pass@1 accuracy under a fixed evaluation protocol.
We also report relative training time reduction, response-length statistics, overlap ratio, top-$p$ acceptance rate, and effective reliable length.
Training time is measured relative to baseline OPD under the same student-teacher pair.

\subsection{Main Results}

\begin{table*}[t]
    % \vspace{3mm}
    \centering
    \small
    \renewcommand\arraystretch{1.2}
    \setlength{\tabcolsep}{5pt}
    \caption{Main results for \method. Accuracy is reported in percent. Time reduction is relative to baseline OPD under the same student-teacher pair.}
    \label{tab:main-results}
    \begin{tabular}{l|c|c|c|c|c|c|c}
       \toprule
       \textbf{Method} & \textbf{AMC23} & \textbf{AIME24} & \textbf{AIME25} & \textbf{HMMT24} & \textbf{HMMT25} & \textbf{Time} & \textbf{Red.} \\
       \midrule
       \rowcolor{gray!10} \multicolumn{8}{c}{\textit{DeepSeek-R1-Distill-Qwen-1.5B / JustRL-DeepSeek-1.5B}} \\
       \opd~& 77.5 & 43.5 & 32.7 & \textbf{23.5} & 18.1 & 6.9h & 0\% \\
       \opd{} (Truncate 4k) & 77.8 & 43.1 & 31.0 & 22.7 & 17.9 & 4.8h & 30.4\% \\
       \rowcolor[HTML]{EDF5F5}\textbf{\method{} (top-$p$)} & 77.3 & 44.0 & 30.8 & 23.3 & \textbf{19.8} & \textbf{4.3h} & \textbf{37.7\%} \\
       \rowcolor[HTML]{D7E8E8}\textbf{\method{} (overlap)} & \textbf{78.4} & \textbf{45.2} & \textbf{33.5} & 22.7 & 18.8 & \textbf{4.1h} & \textbf{40.6\%} \\
       \midrule
       \rowcolor{gray!10} \multicolumn{8}{c}{\textit{DeepSeek-R1-Distill-Qwen-1.5B / DeepSeek-R1-Distill-Qwen-7B}} \\
       \opd~& \textbf{66.4} & \textbf{33.5} & 25.4 & 14.4 & \textbf{15.0} & 12.5h & 0\% \\
       \opd{} (Truncate 4k) & 66.3 & 31.7 & \textbf{25.8} & 14.2 & 14.4 & 6.6h & 47.2\% \\
       \rowcolor[HTML]{EDF5F5}\textbf{\method{} (top-$p$)} & 65.5 & 29.6 & 24.2 & \textbf{14.6} & 14.0 & 4.8h & 61.6\% \\
       \rowcolor[HTML]{D7E8E8}\textbf{\method{} (overlap)} & 65.9 & 33.3 & \textbf{25.8} & \textbf{14.6} & \textbf{15.0} & \textbf{4.0h} & \textbf{68.0\%} \\
       \midrule
       \rowcolor{gray!10} \multicolumn{8}{c}{\textit{Qwen3-1.7B-Base / Qwen3-4B (Non-thinking)}} \\
       \opd~& 28.0 & \textbf{7.3} & 2.9 & 2.1 & 0.6 & 16.5h & 0\% \\
       \opd{} (Truncate 4k) & 28.6 & 6.3 & 2.9 & 2.3 & \textbf{1.0} & 12.9h & 21.8\% \\
       \rowcolor[HTML]{EDF5F5}\textbf{\method{} (top-$p$)} & 28.4 & 5.6 & 2.9 & 1.9 & 0.6 & 11.5h & 30.3\% \\
       \rowcolor[HTML]{D7E8E8}\textbf{\method{} (overlap)} & \textbf{29.1} & \textbf{7.3} & \textbf{4.2} & \textbf{2.9} & \textbf{1.0} & \textbf{10.3h} & \textbf{37.6\%} \\
       \midrule
       \rowcolor{gray!10} \multicolumn{8}{c}{\textit{Qwen3-4B-Base / Qwen3-4B (Non-thinking)}} \\
       \opd~& \textbf{42.7} & 14.8 & 10.6 & 5.4 & 2.9 & 28.5h & 0\% \\
       \opd{} (Truncate 4k) & 41.9 & 15.2 & 13.5 & 4.8 & 2.7 & 22.3h & 21.8\% \\
       \rowcolor[HTML]{EDF5F5}\textbf{\method{} (top-$p$)} & 40.5 & 13.1 & 12.7 & 4.8 & 4.0 & 20.0h & 29.8\% \\
       \rowcolor[HTML]{D7E8E8}\textbf{\method{} (overlap)} & \textbf{42.7} & \textbf{16.3} & \textbf{14.6} & \textbf{6.7} & \textbf{5.6} & \textbf{13.5h} & \textbf{52.6\%} \\
       \bottomrule
    \end{tabular}
    \vspace{-4mm}
\end{table*}

The main result is a consistent efficiency gain without meaningful degradation in downstream accuracy.
Across the main low-compatibility runs, the overlap-ratio variant reduces training time by 35.7\%, 68.0\%, 37.6\%, and 52.6\%, while benchmark accuracy remains close to the corresponding OPD baseline.
Accuracy differences are mixed across individual benchmarks and should be interpreted as secondary: \method~slightly improves several AIME/HMMT scores, but also decreases a few entries such as AIME24 under the DeepSeek-R1-Distill-Qwen-7B teacher and HMMT24 under the JustRL-DeepSeek-1.5B teacher.
This pattern is consistent with the role of \method~as a reliability-aware efficiency mechanism rather than a standalone accuracy-improvement method.
The top-$p$ action-acceptance variant is generally more conservative and often weaker than overlap, suggesting that exact sampled-token acceptability can be too strict for reasoning traces, whereas candidate-space overlap better captures whether the teacher still supplies exploitable local supervision.

\begin{figure*}[t]
\centering
\includegraphics[width=\textwidth]{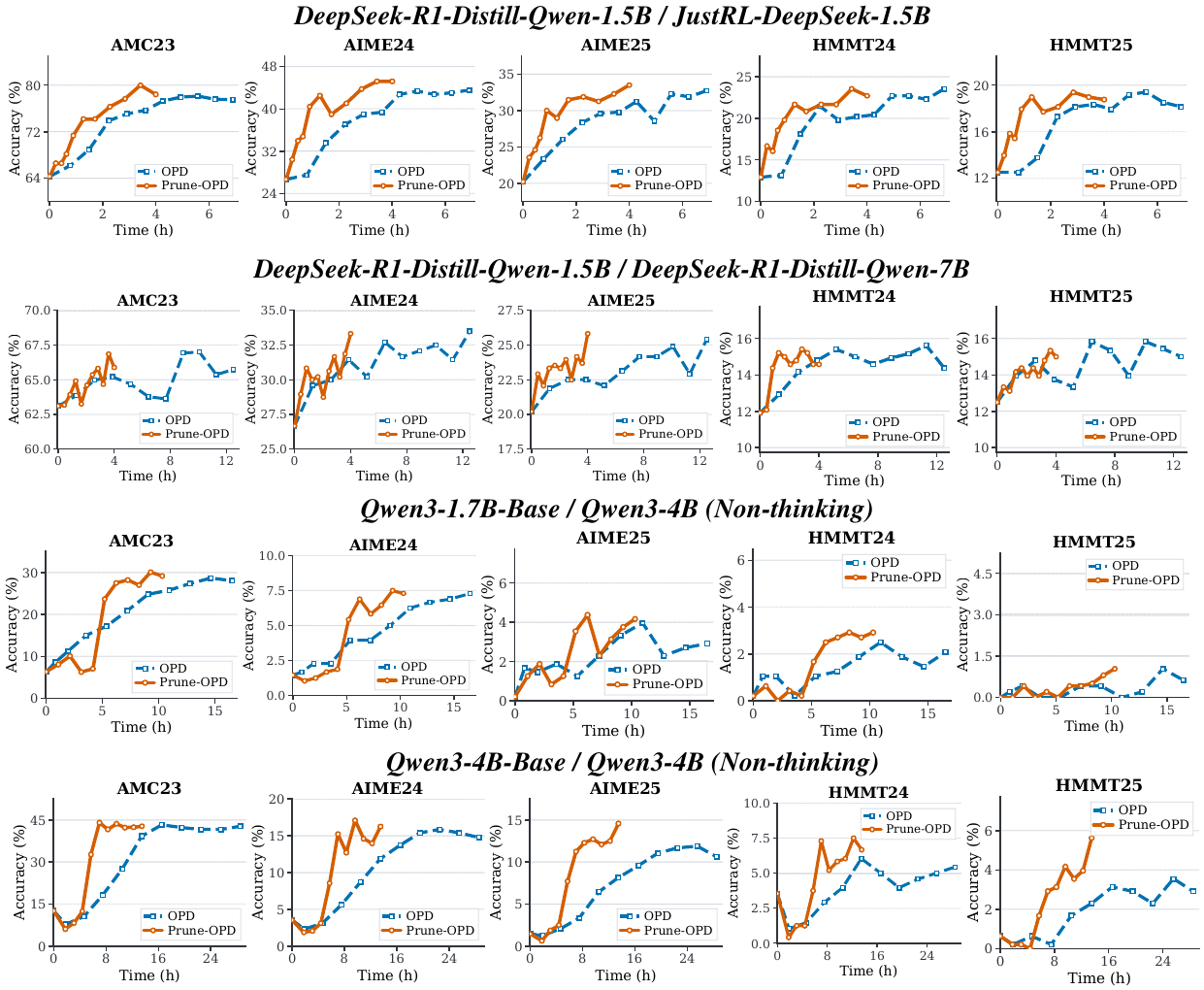}
\vspace{-6mm}
\caption{Accuracy over wall-clock time for the 4 DeepSeek student-teacher pairs. Each panel uses wall-clock time as the x-axis and benchmark accuracy as the y-axis, comparing OPD and \method. A successful curve should match or exceed OPD accuracy while reaching comparable checkpoints earlier in time.}
\label{fig:time-accuracy}
\vspace{-4mm}
\end{figure*}

\subsection{Adaptive Behavior Under High Compatibility}

\begin{figure*}[t]
\vspace{-3mm}
\centering
\includegraphics[width=\textwidth]{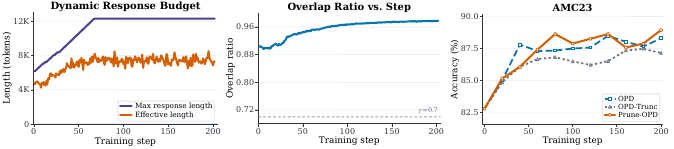}
\vspace{-5mm}
\caption{High-compatibility training dynamics for DeepSeek-R1-Distill-Qwen-7B / Skywork-OR1-7B. Left: effective response length and maximum OPD length versus training step. Middle: overlap ratio versus training step. Right: AMC23 accuracy over training, comparing OPD, \opd{} (Truncate 4k), and \method.}
\label{fig:pair5-high-compatibility-dynamics}
\vspace{-6mm}
\end{figure*}

This setting tests whether \method~uniformly shortens every run.
When the teacher and student have high top-$k$ overlap, the effective-length hit ratio is quickly satisfied and the dynamic controller opens the training window to 12288 tokens.
The resulting training time and accuracy are nearly unchanged from OPD.
Figure~\ref{fig:pair5-high-compatibility-dynamics} should show the corresponding mechanism from three views: the overlap ratio remains high, the dynamic response budget tracks the long reliable context rather than forcing early truncation, and the AMC23 learning curve compares whether \method~matches OPD-level accuracy while avoiding the rigidity of fixed truncation.
This supports the intended interpretation that \method~is an adaptive reliability controller, not a hard length penalty.

\subsection{Training Dynamics}

\begin{figure*}[t]
\centering
\includegraphics[width=\textwidth]{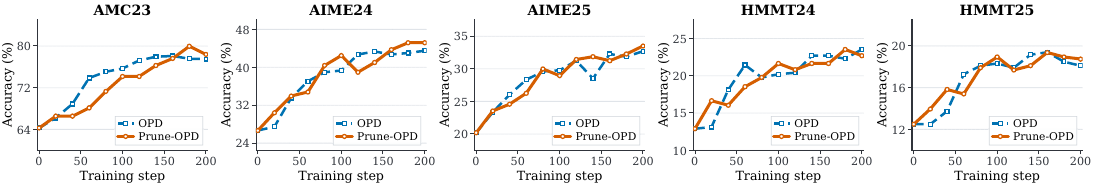}
\vspace{-5mm}
\caption{Training-step accuracy dynamics for DeepSeek-R1-Distill-Qwen-1.5B distilled from JustRL-DeepSeek-1.5B. The five panels report benchmark accuracy over training steps on AMC23, AIME24, AIME25, HMMT24, and HMMT25, comparing OPD and \method.}
\label{fig:pair1-training-accuracy}
\end{figure*}

\begin{figure*}[t]
\vspace{-3mm}
\centering
\includegraphics[width=\textwidth]{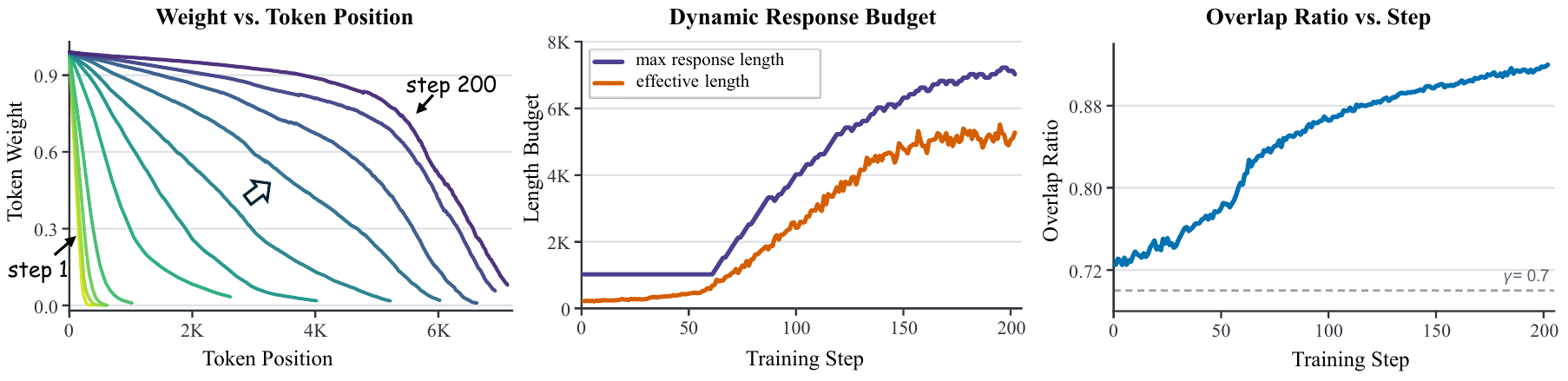}
\vspace{-6mm}
\caption{Training-dynamics diagnostics for DeepSeek-R1-Distill-Qwen-1.5B distilled from JustRL-DeepSeek-1.5B. The panels report mean Prune-OPD weight by token position with curves every 20 training steps from 0 to 200; effective response length and maximum OPD length over training; and overlap ratio over training.}
\label{fig:pair1-training-diagnostics}
\vspace{-4mm}
\end{figure*}

Figure~\ref{fig:pair1-training-accuracy} examines whether \method~changes the learning dynamics or simply removes rollout computation.
For the JustRL-DeepSeek-1.5B teacher, the benchmark-wise step-accuracy curves should show that \method~follows the same learning trajectory as OPD on AMC23, AIME24, AIME25, HMMT24, and HMMT25, rather than achieving speed by sacrificing accuracy.
Figure~\ref{fig:pair1-training-diagnostics} explains where the saved computation comes from: the weight-by-position plot should show suffix reward attenuation after accumulated prefix drift, the effective-length plot should show the dynamic maximum OPD length adapting to the reliable supervision window, and the overlap-ratio curve should confirm that pruning is driven by the measured local compatibility signal.

\begin{table*}[t]
    \centering
    \small
    \renewcommand\arraystretch{1.2}
    \setlength{\tabcolsep}{5pt}
    \caption{High-compatibility behavior. DeepSeek-R1-Distill-Qwen-7B and Skywork-OR1-7B start with overlap ratio around 0.94.}
    \label{tab:high-compatibility}
    \begin{tabular}{l|c|c|c|c|c|c|c}
       \toprule
       \textbf{Method} & \textbf{AMC23} & \textbf{AIME24} & \textbf{AIME25} & \textbf{HMMT24} & \textbf{HMMT25} & \textbf{Time} & \textbf{Red.} \\
       \midrule
       \rowcolor{gray!10} \multicolumn{8}{c}{\textit{DeepSeek-R1-Distill-Qwen-7B / Skywork-OR1-7B}} \\
       \opd~& 88.3 & \textbf{67.1} & \textbf{52.5} & \textbf{33.5} & \textbf{32.1} & 17.5h & 0\% \\
       \opd{} (Truncate 4k) & \textcolor{red!40}{87.2} & \textcolor{red!40}{66.0} & \textbf{\textcolor{red!40}{48.8}} & \textbf{\textcolor{red!40}{31.0}} & \textcolor{red!40}{30.4} & 11.3h & \textbf{35.4\%} \\
       \rowcolor[HTML]{D7E8E8}\textbf{\method{} (overlap)} & \textbf{88.9} & 66.7 & 52.1 & 33.1 & 31.5 & 17.0h & 2.9\% \\
       \bottomrule
    \end{tabular}
\end{table*}

\subsection{Wall-Clock Efficiency}

Figure~\ref{fig:time-accuracy} evaluates the core efficiency claim on the wall-clock axis: \method~should reduce elapsed training time without reducing accuracy, and in some cases can slightly improve it.
If \method~is working as intended, its curves should lie to the left of OPD at comparable accuracy levels, or above OPD at the same wall-clock time.
This presentation directly tests whether \method~reaches OPD-level, and occasionally better, accuracy with less elapsed training time across both DeepSeek teacher-student combinations.
Together with the dynamics in Figures~\ref{fig:pair1-training-accuracy} and~\ref{fig:pair1-training-diagnostics}, these results show that: reliability-aware suffix down-weighting shortens unproductive rollout computation while preserving the useful prefix supervision that drives final performance.

\begin{table}[htbp]
\vspace{-3mm}
\centering
\caption{Overlap-threshold ablation on DeepSeek-R1-Distill-Qwen-1.5B / JustRL-DeepSeek-1.5B. 
% Accuracy is reported in percent.
}
\label{tab:threshold-ablation}
\small
\setlength{\tabcolsep}{4pt}
\renewcommand{\arraystretch}{1.1}
\begin{tabular}{lccccccc}
\toprule
Threshold $\gamma$ & AMC23 & AIME24 & AIME25 & HMMT24 & HMMT25 & Time & Red. \\
\midrule
0.6 & \textbf{78.9} & \textbf{45.4} & 33.1 & 22.0 & \textbf{18.8} & 4.8h & 30.4\%\\
\rowcolor[HTML]{D7E8E8}0.7 & 78.4 & 45.2 & \textbf{33.5} & \textbf{22.7} & \textbf{18.8} & 4.1h & 40.6\% \\
0.8 & 77.8 & 44.5 & 31.5 & 21.9 & 18.0 & 3.1h & 55.1\%\\
0.9 & 76.8 & 43.9 & 30.8 & 21.8 & 17.9 & 2.4h & 65.2\%\\
\bottomrule
\end{tabular}
\vspace{-4mm}
\end{table}

\subsection{Ablation Study}

\noindent\textbf{OPD with simple truncation.}
We include \opd{} (Truncate 4k) as a fixed-budget baseline in Tables~\ref{tab:main-results} and~\ref{tab:high-compatibility}.
This baseline uses the same OPD objective but shortens every rollout uniformly, without observing student-teacher compatibility.
Although simple truncation can reduce cost and may be effective for some teacher-student pairs, it is not a reliable substitute for \method.
In particular, for DeepSeek-R1-Distill-Qwen-1.5B / JustRL-DeepSeek-1.5B and DeepSeek-R1-Distill-Qwen-7B / Skywork-OR1-7B, truncating at a fixed 4K budget degrades performance because it discards useful long-prefix supervision.
This failure mode highlights the limitation of a rigid length rule: it cannot distinguish drifted suffixes from long but still compatible reasoning trajectories.

\noindent\textbf{Pruning metric: overlap ratio vs. top-$p$ acceptance.}
The metric ablation is reported in Table~\ref{tab:main-results}.
Across the completed runs, overlap ratio is the stronger default: it gives equal or larger time reduction than top-$p$ acceptance while generally preserving accuracy better.
We attribute this to its support-level nature: it asks whether student and teacher still share high-probability candidate regions, rather than requiring the sampled student token to fall inside the teacher's visible nucleus.
This makes overlap ratio a smoother reliability signal for long reasoning traces, where multiple locally plausible continuations may remain valid.

% \noindent\textbf{Pruning metric: overlap ratio vs. top-$p$ acceptance.}
% The metric ablation is reported in Table~\ref{tab:main-results}.
% Across the completed runs, overlap ratio is the stronger default: it achieves equal or larger time reduction than top-$p$ acceptance and generally better preserves accuracy.
% The reason is that overlap ratio measures compatibility between the student and teacher high-probability regions, which is exactly the local condition needed for OPD rewards to remain exploitable.
% By contrast, top-$p$ acceptance is an action-level test: it asks whether the sampled student token lies inside the teacher's visible nucleus.
% This can be overly strict for reasoning traces, where multiple locally plausible next tokens may lead to valid continuations even if the exact sampled token is outside the teacher's truncated top-$p$ set.
% Thus, overlap ratio provides a smoother and more robust signal for deciding when to attenuate suffix rewards.

\noindent\textbf{Prune-OPD threshold ablation.}
We ablate the overlap-ratio threshold in Table~\ref{tab:threshold-ablation} on DeepSeek-R1-Distill-Qwen-1.5B / JustRL-DeepSeek-1.5B.
The completed $\gamma=0.7$ setting gives a strong operating point: it preserves OPD-level benchmark performance while reducing training time by 35.7\%.
A lower threshold should be more permissive, producing fewer bad events and longer effective distillation lengths; a higher threshold should prune earlier and may remove useful long reasoning if set too aggressively.
The central conclusion is that $\gamma=0.7$ provides both performance protection and substantial training-efficiency improvement.

\section{Related Work}

\noindent\textbf{On-policy distillation.}
MiniLLM formalized OPD for LLMs under a reverse-KL objective optimized by policy gradient, emphasizing the mode-seeking behavior of reverse KL \citep{gu2024minillm}.
GKD broadened this view by interpolating between on-policy and off-policy data under multiple divergences \citep{agarwal2024policy}.
More recent theory frames OPD as dense KL-constrained RL, where the teacher's per-token log-ratio can be interpreted as an implicit reward and can even be extrapolated beyond the teacher's standard strength \citep{yang2026learning}.
OPD has since appeared in major reasoning and post-training recipes \citep{yang2025qwen3,lu2025onpolicydistillation,zeng2026glm,xiao2026mimo,ko2026scaling,jin2026entropy,jang2026stable,fu2026revisiting,yang2026learning,gu2026coevolvingpolicydistillation}.
It has also been extended to self-distillation and online experiential learning, where a model acts as its own teacher under privileged information, feedback, or sample routing \citep{hubotter2026reinforcement,zhao2026self,he2026far,shenfeld2026self,ye2026policy,sang2026crispcompressedreasoningiterative,kim2026does,ye2026online,yang2026self,li2026unifying,zhao2026selfdistillationmultitokenprediction,ding2026hdpo,song2026ehrag}.
These works establish OPD as a strong post-training primitive.
\method~addresses a complementary reliability problem: even when dense teacher rewards are available on-policy, long rollouts may contain prefixes where the teacher's local signal is no longer exploitable. More discussions of OPD are provided in Appendix~\ref{app:related}

\vspace{-5mm}

\section{Conclusion}
\vspace{-3mm}
OPD provides dense supervision for long reasoning chains; however, when the student's reasoning pattern diverges from that of the teacher, the resulting dense rewards become unreliable.
\method~treats this as a position-wise reliability problem: it checks whether student and teacher remain locally compatible under shared prefix, turns cumulative compatibility failures into a drift signal, rescales OPD rewards, and controls response length using reliable rather than raw token count.
Across diverse distillation settings, overlap-based \method~reduces training time by 37.6\%--68.0\% in low-compatibility regimes while preserving benchmark performance, and it keeps the long training window open when compatibility remains high.
These results show that efficient long-horizon OPD should not be governed by teacher strength or a fixed rollout budget alone.
Instead, dense teacher supervision should be allocated according to its local exploitability on the prefixes that the student actually visits.

\bibliographystyle{plain}

% \bibliography{references}

\begin{thebibliography}{10}

\bibitem{agarwal2024policy}
Rishabh Agarwal, Nino Vieillard, Yongchao Zhou, Piotr Stanczyk, Sabela~Ramos Garea, Matthieu Geist, and Olivier Bachem.
\newblock On-policy distillation of language models: Learning from self-generated mistakes.
\newblock In {\em International Conference on Learning Representations}, 2024.

\bibitem{bengio2015scheduled}
Samy Bengio, Oriol Vinyals, Navdeep Jaitly, and Noam Shazeer.
\newblock Scheduled sampling for sequence prediction with recurrent neural networks.
\newblock {\em Advances in neural information processing systems}, 28, 2015.

\bibitem{busbridge2025distillation}
Dan Busbridge, Amitis Shidani, Floris Weers, Jason Ramapuram, Etai Littwin, and Russell Webb.
\newblock Distillation scaling laws.
\newblock In {\em International Conference on Machine Learning}, pages 5977--6045. PMLR, 2025.

\bibitem{cho2019efficacy}
Jang~Hyun Cho and Bharath Hariharan.
\newblock On the efficacy of knowledge distillation.
\newblock In {\em Proceedings of the IEEE/CVF international conference on computer vision}, pages 4794--4802, 2019.

\bibitem{chung2024scaling}
Hyung~Won Chung, Le~Hou, Shayne Longpre, Barret Zoph, Yi~Tay, William Fedus, Yunxuan Li, Xuezhi Wang, Mostafa Dehghani, Siddhartha Brahma, et~al.
\newblock Scaling instruction-finetuned language models.
\newblock {\em Journal of Machine Learning Research}, 25(70):1--53, 2024.

\bibitem{ding2026hdpo}
Ken Ding.
\newblock Hdpo: Hybrid distillation policy optimization via privileged self-distillation, 2026.

\bibitem{fu2026revisiting}
Yuqian Fu, Haohuan Huang, Kaiwen Jiang, Yuanheng Zhu, and Dongbin Zhao.
\newblock Revisiting on-policy distillation: Empirical failure modes and simple fixes, 2026.

\bibitem{gu2026coevolvingpolicydistillation}
Naibin Gu, Chenxu Yang, Qingyi Si, Chuanyu Qin, Dingyu Yao, Peng Fu, Zheng Lin, Weiping Wang, Nan Duan, and Jiaqi Wang.
\newblock Co-evolving policy distillation, 2026.

\bibitem{gu2024minillm}
Yuxian Gu, Li~Dong, Furu Wei, and Minlie Huang.
\newblock Minillm: Knowledge distillation of large language models.
\newblock In {\em The twelfth international conference on learning representations}, 2024.

\bibitem{guo2025deepseek}
Daya Guo, Dejian Yang, Haowei Zhang, Junxiao Song, Peiyi Wang, Qihao Zhu, Runxin Xu, Ruoyu Zhang, Shirong Ma, Xiao Bi, et~al.
\newblock Deepseek-r1 incentivizes reasoning in llms through reinforcement learning.
\newblock {\em Nature}, 645(8081):633--638, 2025.

\bibitem{he2025justrl}
Bingxiang He, Zekai Qu, Zeyuan Liu, Yinghao Chen, Yuxin Zuo, Cheng Qian, Kaiyan Zhang, Weize Chen, Chaojun Xiao, Ganqu Cui, et~al.
\newblock Justrl: Scaling a 1.5b llm with a simple rl recipe, 2025.

\bibitem{he2026far}
Bingxiang He, Yuxin Zuo, Zeyuan Liu, Shangziqi Zhao, Zixuan Fu, Junlin Yang, Cheng Qian, Kaiyan Zhang, Yuchen Fan, Ganqu Cui, et~al.
\newblock How far can unsupervised rlvr scale llm training?, 2026.

\bibitem{hinton2015distilling}
Geoffrey Hinton, Oriol Vinyals, and Jeff Dean.
\newblock Distilling the knowledge in a neural network.
\newblock {\em arXiv preprint arXiv:1503.02531}, 2015.

\bibitem{hubotter2026reinforcement}
Jonas Hubotter, Frederike Lubeck, Lejs Behric, Anton Baumann, Marco Bagatella, Daniel Marta, Ido Hakimi, Idan Shenfeld, Thomas~Kleine Buening, Carlos Guestrin, et~al.
\newblock Reinforcement learning via self-distillation, 2026.

\bibitem{jang2026stable}
Ijun Jang, Jewon Yeom, Juan Yeo, Hyunggu Lim, and Taesup Kim.
\newblock Stable on-policy distillation through adaptive target reformulation, 2026.

\bibitem{jiao2020tinybert}
Xiaoqi Jiao, Yichun Yin, Lifeng Shang, Xin Jiang, Xiao Chen, Linlin Li, Fang Wang, and Qun Liu.
\newblock Tinybert: Distilling bert for natural language understanding.
\newblock In {\em Findings of the Association for Computational Linguistics: EMNLP 2020}, pages 4163--4174, 2020.

\bibitem{jin2026entropy}
Woogyeol Jin, Taywon Min, Yongjin Yang, Swanand~Ravindra Kadhe, Yi~Zhou, Dennis Wei, Nathalie Baracaldo, and Kimin Lee.
\newblock Entropy-aware on-policy distillation of language models, 2026.

\bibitem{kim2026does}
Jeonghye Kim, Xufang Luo, Minbeom Kim, Sangmook Lee, Dohyung Kim, Jiwon Jeon, Dongsheng Li, and Yuqing Yang.
\newblock Why does self-distillation (sometimes) degrade the reasoning capability of llms?, 2026.

\bibitem{kim2016sequence}
Yoon Kim and Alexander~M. Rush.
\newblock Sequence-level knowledge distillation.
\newblock In {\em Proceedings of the 2016 Conference on Empirical Methods in Natural Language Processing}, pages 1317--1327, 2016.

\bibitem{ko2026scaling}
Jongwoo Ko, Sara Abdali, Young~Jin Kim, Tianyi Chen, and Pashmina Cameron.
\newblock Scaling reasoning efficiently via relaxed on-policy distillation, 2026.

\bibitem{li2026unifying}
Gengsheng Li, Tianyu Yang, Junfeng Fang, Mingyang Song, Mao Zheng, Haiyun Guo, Dan Zhang, Jinqiao Wang, and Tat-Seng Chua.
\newblock Unifying group-relative and self-distillation policy optimization via sample routing, 2026.

\bibitem{li2026rethinking}
Yaxuan Li, Yuxin Zuo, Bingxiang He, Jinqian Zhang, Chaojun Xiao, Cheng Qian, Tianyu Yu, Huan-ang Gao, Wenkai Yang, Zhiyuan Liu, et~al.
\newblock Rethinking on-policy distillation of large language models: Phenomenology, mechanism, and recipe.
\newblock {\em arXiv preprint arXiv:2604.13016}, 2026.

\bibitem{li2025small}
Yuetai Li, Xiang Yue, Zhangchen Xu, Fengqing Jiang, Luyao Niu, Bill~Yuchen Lin, Bhaskar Ramasubramanian, and Radha Poovendran.
\newblock Small models struggle to learn from strong reasoners.
\newblock In {\em Findings of the Association for Computational Linguistics: ACL 2025}, pages 25366--25394, 2025.

\bibitem{lu2025onpolicydistillation}
Kevin Lu and Thinking~Machines Lab.
\newblock On-policy distillation.
\newblock {\em Thinking Machines Lab: Connectionism}, 2025.
\newblock https://thinkingmachines.ai/blog/on-policy-distillation.

\bibitem{mirzadeh2020improved}
Seyed-Iman Mirzadeh, Mehrdad Farajtabar, Ang Li, Nir Levine, Akihiro Matsukawa, and Hassan Ghasemzadeh.
\newblock Improved knowledge distillation via teacher assistant.
\newblock In {\em Proceedings of the AAAI Conference on Artificial Intelligence}, volume~34, pages 5191--5198, 2020.

\bibitem{sang2026crispcompressedreasoningiterative}
Hejian Sang, Yuanda Xu, Zhengze Zhou, Ran He, Zhipeng Wang, and Jiachen Sun.
\newblock Crisp: Compressed reasoning via iterative self-policy distillation, 2026.

\bibitem{sanh2019distilbert}
Victor Sanh, Lysandre Debut, Julien Chaumond, and Thomas Wolf.
\newblock Distilbert, a distilled version of bert: Smaller, faster, cheaper and lighter, 2019.

\bibitem{sanh2021multitask}
Victor Sanh, Albert Webson, Colin Raffel, Stephen Bach, Lintang Sutawika, Zaid Alyafeai, Antoine Chaffin, Arnaud Stiegler, Arun Raja, Manan Dey, et~al.
\newblock Multitask prompted training enables zero-shot task generalization.
\newblock In {\em International Conference on Learning Representations}, 2021.

\bibitem{shao2024deepseekmath}
Zhihong Shao, Peiyi Wang, Qihao Zhu, Runxin Xu, Junxiao Song, Xiao Bi, Haowei Zhang, Mingchuan Zhang, Y.~K. Li, Yang Wu, et~al.
\newblock Deepseekmath: Pushing the limits of mathematical reasoning in open language models.
\newblock {\em arXiv preprint arXiv:2402.03300}, 2024.

\bibitem{shenfeld2026self}
Idan Shenfeld, Mehul Damani, Jonas Hubotter, and Pulkit Agrawal.
\newblock Self-distillation enables continual learning, 2026.

\bibitem{song2026ehrag}
Yifan Song, Xingjian Tao, Zhicheng Yang, Yihong Luo, and Jing Tang.
\newblock Ehrag: Bridging semantic gaps in lightweight graphrag via hybrid hypergraph construction and retrieval.
\newblock {\em arXiv preprint arXiv:2604.17458}, 2026.

\bibitem{wang2020minilm}
Wenhui Wang, Furu Wei, Li~Dong, Hangbo Bao, Nan Yang, and Ming Zhou.
\newblock Minilm: Deep self-attention distillation for task-agnostic compression of pre-trained transformers.
\newblock {\em Advances in Neural Information Processing Systems}, 33:5776--5788, 2020.

\bibitem{wei2021finetuned}
Jason Wei, Maarten Bosma, Vincent Zhao, Kelvin Guu, Adams~Wei Yu, Brian Lester, Nan Du, Andrew~M. Dai, and Quoc~V. Le.
\newblock Finetuned language models are zero-shot learners.
\newblock In {\em International Conference on Learning Representations}, 2021.

\bibitem{xiao2026mimo}
Bangjun Xiao, Bingquan Xia, Bo~Yang, Bofei Gao, Bowen Shen, Chen Zhang, Chenhong He, Chiheng Lou, Fuli Luo, Gang Wang, et~al.
\newblock Mimo-v2-flash technical report, 2026.

\bibitem{xu2026envfactoryscalingtooluseagents}
Minrui Xu, Zilin Wang, Mengyi DENG, Zhiwei Li, Zhicheng Yang, Xiao Zhu, Yinhong Liu, Boyu Zhu, Baiyu Huang, Chao Chen, Heyuan Deng, Fei Mi, Lifeng Shang, Xingshan Zeng, and Zhijiang Guo.
\newblock Envfactory: Scaling tool-use agents via executable environments synthesis and robust rl, 2026.

\bibitem{yang2025qwen3}
An~Yang, Anfeng Li, Baosong Yang, Beichen Zhang, Binyuan Hui, Bo~Zheng, Bowen Yu, Chang Gao, Chengen Huang, Chenxu Lv, et~al.
\newblock Qwen3 technical report, 2025.

\bibitem{yang2026self}
Chenxu Yang, Chuanyu Qin, Qingyi Si, Minghui Chen, Naibin Gu, Dingyu Yao, Zheng Lin, Weiping Wang, Jiaqi Wang, and Nan Duan.
\newblock Self-distilled rlvr, 2026.

\bibitem{yang2026learning}
Wenkai Yang, Weijie Liu, Ruobing Xie, Kai Yang, Saiyong Yang, and Yankai Lin.
\newblock Learning beyond teacher: Generalized on-policy distillation with reward extrapolation, 2026.

\bibitem{accordion}
Zhicheng Yang, Zhijiang Guo, Yinya Huang, Yongxin Wang, Wenlei Shi, Yiwei Wang, Xiaodan Liang, and Jing Tang.
\newblock Accordion-thinking: Self-regulated step summaries for efficient and readable llm reasoning.
\newblock In {\em Forty-Third International Conference on Machine Learning}, 2026.

\bibitem{dars}
Zhicheng Yang, Zhijiang Guo, Yinya Huang, Yongxin Wang, Dongchun Xie, Hanhui Li, Yiwei Wang, Xiaodan Liang, and Jing Tang.
\newblock Depth-breadth synergy in rlvr: Unlocking llm reasoning gains with adaptive exploration.
\newblock In {\em Forty-Third International Conference on Machine Learning}, 2026.

\bibitem{yang2025optibench}
Zhicheng Yang, Yiwei Wang, Yinya Huang, Zhijiang Guo, Wei Shi, Xiongwei Han, Liang Feng, Linqi Song, Xiaodan Liang, and Jing Tang.
\newblock Optibench meets resocratic: Measure and improve {LLM}s for optimization modeling.
\newblock In {\em The Thirteenth International Conference on Learning Representations}, 2025.

\bibitem{ye2026online}
Tianzhu Ye, Li~Dong, Qingxiu Dong, Xun Wu, Shaohan Huang, and Furu Wei.
\newblock Online experiential learning for language models, 2026.

\bibitem{ye2026policy}
Tianzhu Ye, Li~Dong, Xun Wu, Shaohan Huang, and Furu Wei.
\newblock On-policy context distillation for language models, 2026.

\bibitem{yu2025dapo}
Qiying Yu, Zheng Zhang, Ruofei Zhu, Yufeng Yuan, Xiaochen Zuo, Yu~Yue, Weinan Dai, Tiantian Fan, Gaohong Liu, Lingjun Liu, et~al.
\newblock Dapo: An open-source llm reinforcement learning system at scale, 2025.

\bibitem{zeng2026glm}
Aohan Zeng, Xin Lv, Zhenyu Hou, Zhengxiao Du, Qinkai Zheng, Bin Chen, Da~Yin, Chendi Ge, Chengxing Xie, Cunxiang Wang, et~al.
\newblock Glm-5: From vibe coding to agentic engineering, 2026.

\bibitem{zhao2026selfdistillationmultitokenprediction}
Guoliang Zhao, Ruobing Xie, An~Wang, Shuaipeng Li, Huaibing Xie, and Xingwu Sun.
\newblock Self-distillation for multi-token prediction, 2026.

\bibitem{zhao2026self}
Siyan Zhao, Zhihui Xie, Mengchen Liu, Jing Huang, Guan Pang, Feiyu Chen, and Aditya Grover.
\newblock Self-distilled reasoner: On-policy self-distillation for large language models, 2026.

\end{thebibliography}

\appendix
\section{Appendix}
\subsection{Broader Impact}
\label{app:impact}
This research contributes to the development of more efficient and reliable training protocols for LLMs specializing in complex reasoning. By introducing \method, we provide a mechanism to mitigate the risks of ``reward hacking'' and training instability inherent in dense on-policy distillation, where models might otherwise learn from low-quality or drifted teacher signals. The significant reduction in training compute (up to 68\%) directly addresses the environmental concerns associated with the high energy consumption of LLM post-training. By making sophisticated reasoning models more accessible to train with fewer resources, this work democratizes the ability to develop high-performing AI, potentially accelerating progress in scientific discovery.

Despite these benefits, we recognize that enhancing the reasoning capabilities of LLMs could inadvertently lower the barrier for generating sophisticated misinformation or dual-use content. While \method focuses on the process of learning rather than the content of the data, the resulting models could be more persuasive or harder to detect if used for deceptive purposes. To mitigate these risks, we emphasize the importance of using \method in conjunction with safety-aligned teacher models and rigorous verifiable RL frameworks. We are committed to transparency and reproducibility by providing detailed experimental setups and hyperparameter configurations, enabling the community to audit the reliability of these training dynamics and ensure that reasoning improvements do not come at the cost of model safety or ethical alignment.

\subsection{LLM Usage Declaration}
\label{app:llm_usage}
This manuscript uses LLMs strictly for the purpose of language editing and textual polishing to enhance presentation quality. We declare that the novel ideas, methodological framework, experimental execution, and data analysis are the original work of the authors. All content modified by AI tools has been carefully reviewed and validated by the authors to ensure accuracy.

\subsection{Limitations}

\method~relies on top-$k$ local-distribution statistics.
The primary overlap metric can miss cases where student and teacher share candidate tokens but rank or score them very differently, while the top-$p$ action-acceptance variant can be too strict for stylistically different but valid reasoning paths.
The current decay is linear in cumulative bad events and assumes that local compatibility correlates with reward exploitability; this can fail when a low-overlap prefix later returns to a teacher-compatible state.
The implementation also only rescales OPD rewards and does not attach a GRPO fallback to zero-reliability tokens.
Finally, the current experiments are limited to math reasoning with DeepSeek, Qwen, and Skywork-style models; agentic and multi-turn settings remain future work.

\subsection{Future Work}

A natural extension is to combine \method~with GRPO in a reliability-gated objective.
Before the reliability weight decays to zero, the model would still optimize the OPD loss and benefit from dense teacher guidance on locally compatible prefixes.
After the reliability weight reaches zero, the unreliable suffix would switch to a GRPO loss, allowing the student to continue exploring with outcome-level feedback instead of forcing it to follow a teacher distribution that is no longer locally exploitable.
Such a hybrid objective could combine the sample efficiency of teacher-guided distillation with the exploratory benefits of reinforcement learning, and may further improve long-horizon reasoning beyond reliability-aware truncation alone.

\subsection{Additional Related Work}
\label{app:related}
\paragraph{Knowledge distillation and off-policy distillation.}
Knowledge distillation transfers information from a teacher to a student by matching the teacher's soft output distribution \citep{hinton2015distilling}.
For autoregressive sequence models, sequence-level distillation trains the student on teacher-generated outputs \citep{kim2016sequence}, and this off-policy paradigm underlies many compact language-model distillation methods \citep{sanh2019distilbert,jiao2020tinybert,wang2020minilm,gu2026coevolvingpolicydistillation}.
Supervised fine-tuning and instruction tuning similarly improve downstream behavior by training on curated responses or demonstrations \citep{chung2024scaling,sanh2021multitask,wei2021finetuned}.
The common limitation is train-inference distribution mismatch: the student is optimized on teacher or reference trajectories, but at inference time it must condition on prefixes sampled from its own policy, an exposure-bias problem that compounds over long generations \citep{bengio2015scheduled}.
\method~follows the OPD motivation that distillation should be computed on student-visited states, but asks a further question: once the student is on-policy, which visited states still admit reliable dense teacher supervision?

\paragraph{OPD dynamics and mechanisms.}
Recent OPD analysis shows that OPD success is governed by thinking-pattern consistency and by whether the teacher contributes new knowledge beyond what the student has already acquired \citep{li2026rethinking}.
At the token level, successful OPD exhibits progressive alignment on high-probability tokens: top-$k$ overlap rises, overlap-token advantage improves, and the entropy gap narrows.
The same analysis further shows that reward quality degrades with trajectory depth, and that a reward can remain globally correlated with final correctness while failing to provide locally useful gradients around the student's current policy.
\method~directly operationalizes these diagnostic findings.
Instead of treating overlap ratio, entropy gap, or teacher acceptability as post-hoc measurements, it turns per-prefix compatibility into an online reliability signal that attenuates future OPD rewards and controls rollout length.

\paragraph{Capacity gap and distillability.}
A separate line of work studies when a teacher is too strong, too different, or too complex for the student to imitate.
% \citet{cho2019efficacy} show that overly capable teachers can hurt student performance, and \citet{mirzadeh2020improved} introduce teacher assistants to bridge large capacity gaps.
Studies have shown that overly capable teachers can hurt student performance \citep{cho2019efficacy}, and teacher assistants have been introduced to bridge large capacity gaps \citep{mirzadeh2020improved}.
Distillation scaling laws identify non-monotonic interactions among teacher quality, student size, and data volume \citep{busbridge2025distillation}.
For reasoning models, small students can struggle to learn from strong reasoners with long chain-of-thought traces, suggesting a learnability gap in distilling complex reasoning behavior \citep{li2025small}.
These results caution against assuming that teacher benchmark strength alone determines distillation quality.
\method~studies the same issue from an on-policy, position-wise perspective: a teacher may be useful globally but unreliable at specific depths of a student-generated trajectory.

\noindent\textbf{Reasoning RL and long-horizon supervision.}
Modern reasoning models rely on verifiable RL, long chain-of-thought rollouts, and large mathematical training sets \citep{shao2024deepseekmath,guo2025deepseek,yu2025dapo,he2025justrl,accordion,dars,yang2025optibench,xu2026envfactoryscalingtooluseagents}.
Long responses create room for search and self-correction, but they also make dense supervision expensive and harder to trust.
Prior OPD results identify a response-length sweet spot: short responses provide too little dense signal, while very long responses can trigger late-stage collapse and suffix-to-prefix instability \citep{li2026rethinking}.
\method~is designed for this long-horizon setting.
It does not impose a fixed response budget; it estimates which prefixes still carry reliable teacher signal and uses that estimate for both reward scaling and dynamic response-length control.

\subsection{Reliability Rationale}
\label{app:rationale}

\method~is motivated by the distinction between globally informative teacher rewards and locally exploitable token-level supervision on student-visited prefixes~\citep{li2026rethinking}.
Overlap ratio and related compatibility metrics are therefore used online: repeated low-compatibility events are treated as evidence that the student trajectory has drifted into a region where uniform dense OPD rewards are less trustworthy.
The method does not assume that every low-overlap token is wrong; it only assumes that cumulative compatibility failure is a useful path-dependent signal for attenuating later suffix rewards.

The monotone reliability decay is a deliberately conservative design.
Later tokens can become locally compatible because the teacher adapts to an already drifted prefix, not because the full reasoning trajectory has returned to a teacher-compatible state.
Using cumulative decay avoids re-amplifying suffix rewards after repeated drift events, while the base weight $w_{\mathrm{base}}$ separates the reliability estimate $R_{\tau}$ from the optimization scale $L_{\tau}$.
This lets \method~down-weight unreliable suffix supervision without turning the method into a brittle hard filter.

\subsection{Implementation Details}

\method~is implemented as a post-processing step on the \opd~reward tensor.
The baseline reward tensor has shape $[B,T,k]$, where $B$ is batch size, $T$ is response length, and $k$ is the top-$k$ token dimension.
For each valid response position, the method computes the scalar loss weight $L_{\tau}$ and multiplies every candidate reward at that position by the same scalar:
\begin{equation}
    \widetilde{r}_{b,\tau,j}=L_{b,\tau}r_{b,\tau,j}.
\end{equation}
Padding positions have $R_{\tau}=0$ and $L_{\tau}=0$.
The advantage estimator remains \texttt{token\_reward\_direct}, so the scaled reward is used directly as the token-level advantage.

The current code path keeps the baseline isolated.
When \texttt{prune\_opd.enable=False}, the original \opd/\grpo behavior is unchanged.
When enabled, the main experiments use \texttt{overlap\_ratio}; \texttt{teacher\_top\_p\_accept} is retained as an action-level variant.
The teacher top-$k$ size is controlled by \texttt{actor\_rollout\_ref.rollout.log\_prob\_top\_k} and is set to 16 in the current experiments.

\subsection{Prompt Template}
\label{app:prompt-template}

We use the same DAPO-Math prompt template as the OPD dynamics study~\citep{li2026rethinking}.
For each math problem, the raw question is inserted into the following instruction.

\definecolor{promptboxbg}{HTML}{F2E9FF}
\definecolor{promptboxframe}{HTML}{7E57C2}
\begin{center}
\setlength{\fboxsep}{8pt}
\fcolorbox{promptboxframe}{promptboxbg}{%
\begin{minipage}{0.92\linewidth}
\textbf{DAPO-Math Prompt Template}\\[0.35em]
\small
\{Question\} Please reason step by step, and put your final answer within \textbackslash boxed\{\}.
\end{minipage}}
\end{center}

\subsection{Primary Configurations}
Table~\ref{tab:hyperparams} summarizes the primary hyperparameters used across our \method~experiments.
Unless otherwise stated, all benchmark performance reported in the paper is Avg@16 accuracy: for each benchmark problem, evaluation uses 16 sampled responses under the fixed evaluation protocol and reports the average correctness across those samples. For high compatibility pair (DeepSeek-R1-Distill-Qwen-7B / Skywork-OR1-7), the init dynamic length is set as 6144; For other pairs, the init dynamic length is set as 1024.

\begin{table}[h]
\centering
\caption{\textbf{Default hyperparameters for \method.}}
\label{tab:hyperparams}
\small
\begin{tabular}{lc}
\toprule
\textbf{Item} & \textbf{Value} \\
\midrule
Training dataset & DAPO-Math-17K \\
Training temperature & 1.0 \\
Teacher temperature & 1.0 \\
Rollout number & 4 \\
Mini batch size & 64 \\
LogProb top-$K$ & 16 \\
Top-$K$ strategy & Student Top-$K$ \\
Reward weight mode & Student probability \\
Max prompt length & 1,024 \\
Max response length & 8,192--12,288 \\
Validation max response length & 31,744 \\
Learning rate & $1\times10^{-6}$ \\
Training steps & 203 \\
KL coefficient & 0.0 \\
Loss aggregation & \texttt{token-mean} \\
Overlap threshold $\gamma$ & 0.7 \\
Top-$p$ threshold $p$ & 0.95 \\
$w_{\mathrm{drop}}$ & 0.01 \\
$w_{\mathrm{base}}$ & 0.5 \\
Dynamic length init/min/max & 1024 / 1024 / 12288 \\
Dynamic length step & 100 \\
Hit ratio threshold & 0.1 \\
Length margin & 100 \\
Shrink patience & 3 \\
\bottomrule
\end{tabular}
\end{table}

\subsection{Top-$p$ Action-Acceptance Metric}
\label{app:top-p-metric}

In addition to overlap ratio, we evaluate a stricter action-level compatibility metric.
At position $\tau$, the student samples the actual next token $y_{\tau}$, while the teacher returns top-$k$ tokens and probabilities under the same prefix:
\begin{equation}
    \topk = \{v_{\tau,1}, \ldots, v_{\tau,k}\}, \qquad
    q_{\tau,1} \geq q_{\tau,2} \geq \cdots \geq q_{\tau,k}.
\end{equation}
Given a top-$p$ threshold $p$, define the cumulative visible mass
\begin{equation}
    S_{\tau,m} = \sum_{j=1}^{m} q_{\tau,j}.
\end{equation}
If a smallest $m_{\tau}$ satisfies $S_{\tau,m_{\tau}}\geq p$, the visible nucleus is
\begin{equation}
    \accept(p,k)=\{v_{\tau,1},\ldots,v_{\tau,m_{\tau}}\}.
\end{equation}
If the returned top-$k$ mass is still below $p$, the true nucleus extends beyond the observed set, so we conservatively use $\accept(p,k)=\topk$.
The bad event is $B_{\tau}=\mathbf{1}[y_{\tau}\notin\accept(p,k)]$.
In the reported top-$p$ runs, $p=0.95$.

\subsection{Why \method Can Improve Accuracy for Low-Overlap Qwen Pairs?}
\label{app:qwen-low-overlap}

\begin{figure*}[t]
\centering
\includegraphics[width=\textwidth]{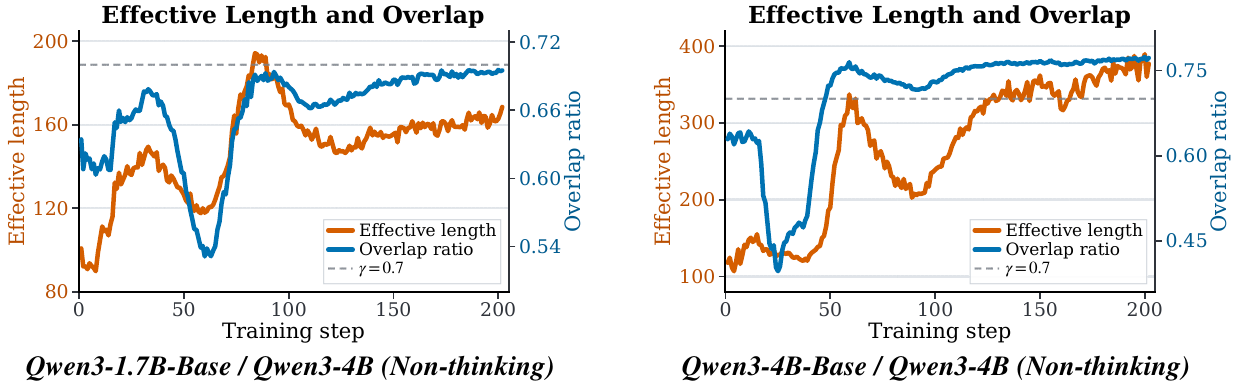}
\caption{Short effective OPD windows in the low-overlap Qwen3 distillation pairs. For Qwen3-1.7B-Base / Qwen3-4B (Non-thinking) and Qwen3-4B-Base / Qwen3-4B (Non-thinking), low overlap causes \method~to concentrate OPD supervision within a few hundred reliable tokens, whereas the OPD baseline keeps training on responses up to 12,288 tokens.}
\label{fig:appendix-qwen-short-effective-length}
\end{figure*}

Table~\ref{tab:main-results} shows that the two Qwen3 teacher-student pairs, Qwen3-1.7B-Base / Qwen3-4B (Non-thinking) and Qwen3-4B-Base / Qwen3-4B (Non-thinking), are the clearest cases where \method~improves both training efficiency and benchmark accuracy.
We interpret this as a low-overlap regime where reliability-aware pruning acts not only as a compute-saving mechanism, but also as a gradient denoising mechanism.
As shown in Figure~\ref{fig:appendix-qwen-short-effective-length}, these pairs exhibit low student-teacher overlap, and \method~therefore keeps the effective OPD length at only a few hundred tokens.
In contrast, the corresponding OPD baseline uses a maximum response budget of 12,288 tokens.

This mismatch creates a plausible failure mode for unpruned OPD.
Only the early prefix contains locally exploitable teacher supervision, while the long suffix contributes a large number of token-level gradients produced on drifted, low-overlap prefixes.
Because those suffix tokens vastly outnumber the useful prefix tokens, their gradients can dominate the update even if each individual suffix reward is weak or noisy.
\method~removes this imbalance by attenuating and truncating low-reliability suffixes, concentrating optimization on the short prefix where the teacher remains actionable.
This explains why these two Qwen3 pairs can gain accuracy in addition to saving time.
For the other teacher-student combinations, compatibility is higher or the useful supervision window is already long enough that pruning mainly reduces computation while preserving accuracy, rather than substantially changing the effective training signal.

\subsection{Additional Diagnostics}

\begin{figure*}[t]
\centering
\includegraphics[width=\textwidth]{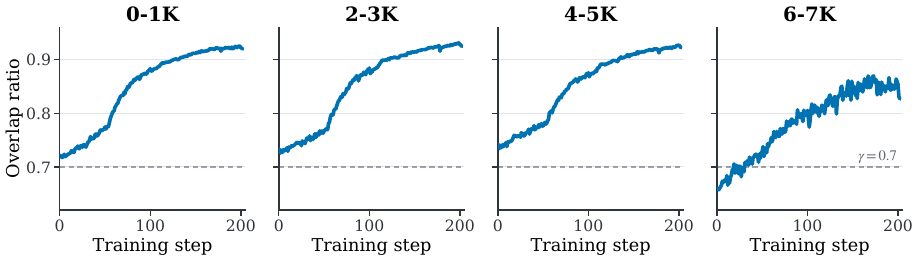}
\caption{OPD baseline overlap-ratio training dynamics for DeepSeek-R1-Distill-Qwen-1.5B / JustRL-DeepSeek-1.5B. Each panel plots overlap ratio versus training step for a token-position band: 0--1K, 2--3K, 4--6K, and 7--8K. This diagnostic shows how local student-teacher compatibility evolves at different trajectory depths under unpruned OPD.}
\label{fig:appendix-opd-overlap-dynamics}
\end{figure*}

\begin{figure*}[t]
\centering
\includegraphics[width=\textwidth]{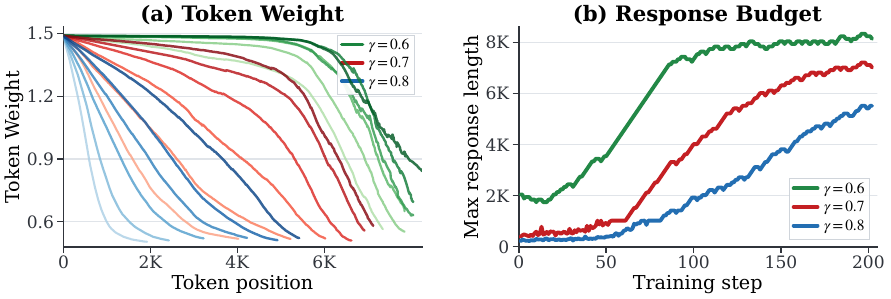}
\caption{Prune-OPD threshold diagnostics for DeepSeek-R1-Distill-Qwen-1.5B / JustRL-DeepSeek-1.5B. Left: mean Prune-OPD weight as a function of token position under three overlap thresholds, $\gamma=0.6,0.7,0.8$; for each threshold, the curves are taken at training steps 100, 120, 140, 160, 180, and 200. Right: maximum OPD response length over training steps under the same thresholds. Together, these diagnostics show how the reliability threshold changes suffix down-weighting and dynamic rollout budgets.}
\label{fig:appendix-threshold-diagnostics}
\end{figure*}

The appendix diagnostics focus on the compatibility signals that motivate \method.
Figure~\ref{fig:appendix-opd-overlap-dynamics} reports how overlap ratio evolves across depth bands under unpruned OPD, and Figure~\ref{fig:appendix-threshold-diagnostics} shows how the overlap threshold controls suffix down-weighting and the dynamic response budget.

\paragraph{Depth-wise overlap under OPD.}
Figure~\ref{fig:appendix-opd-overlap-dynamics} isolates the compatibility dynamics of the unpruned OPD baseline.
The shallow 0--1K band acts as a reference for early-prefix supervision, while the later bands test whether teacher supervision remains locally aligned after long student-generated reasoning prefixes.
The key observation is that compatibility is depth-dependent: later token-position bands are the regime where overlap is most fragile and where uniform dense rewards are least likely to provide useful local gradients.
This supports the central premise of \method: the inefficiency of long-horizon OPD is not only that suffixes are expensive, but that the suffix rewards are often produced on prefixes where the teacher and student have already drifted apart.

\paragraph{Threshold sensitivity.}
Figure~\ref{fig:appendix-threshold-diagnostics} explains how the reliability threshold changes the intervention.
A larger threshold declares compatibility failures earlier, which makes the token-level weight decay sooner and yields a shorter dynamic OPD budget; a smaller threshold is more permissive and preserves longer suffix supervision.
The main setting $\gamma=0.7$ lies between these extremes: it removes a substantial amount of low-reliability suffix computation while avoiding the overly rigid behavior of a hard fixed-length truncation.
Thus the threshold primarily controls where \method~sits on the reliability--compute axis, rather than changing the underlying OPD objective.

\subsection{Compute Reporting}
\label{app:compute-reporting}

The dominant compute cost comes of our work from post-training student models with OPD-style teacher scoring and from benchmark evaluation.
For each teacher-student pair, OPD, fixed truncation, and \method~are run under the same training stack and matched hardware allocation, so the reported wall-clock comparisons isolate the effect of reliability-aware pruning rather than changes in infrastructure.
The main compute evidence is therefore reported as elapsed training time in Tables~\ref{tab:main-results}, \ref{tab:high-compatibility}, and \ref{tab:threshold-ablation}, and in Figure~\ref{fig:time-accuracy}

\end{document}